\documentclass[conference]{IEEEtran}
\IEEEoverridecommandlockouts
\usepackage{hyperref}
\usepackage[table]{xcolor} 
\usepackage{graphicx}
\usepackage{textcomp}
\usepackage{wrapfig}
\usepackage{multirow}
\usepackage[numbers,sort&compress]{natbib}
\usepackage{amsmath,amssymb,amsfonts}
\usepackage{amsthm}
\usepackage{mathrsfs}
\usepackage{manyfoot}
\usepackage{booktabs}
\usepackage{algpseudocode}
\usepackage{listings}
\usepackage{subcaption}  
\newcommand{\tperc}{\scalebox{0.7}{\%}}
\definecolor{lightyellow}{rgb}{1,1,0.6}
\definecolor{lightorange}{rgb}{1,0.8,0.8}

\begin{document}
\title{Latent Representation Learning for Multimodal Brain Activity Translation}
\author{
    \vspace{-.5cm}
    \IEEEauthorblockA{  
        \IEEEauthorblockN{
            Arman Afrasiyabi$^{1,4,9}$, Dhananjay Bhaskar$^{1,4,9}$, Erica L. Busch$^{3,8,9}$, Laurent Caplette$^{3,8,9}$, Rahul Singh$^{1,8,9}$,
        }  
        \IEEEauthorblockN{
            Guillaume Lajoie$^{5,6,7}$, Nicholas B. Turk-Browne$^{3,8,9,*}$, Smita Krishnaswamy$^{1,4,2,8,9,*}$ 
        }
        \IEEEauthorblockN{
            \small  Department of \{$^1$Computer Science, $^2$Applied Mathematics, $^3$Psychology, $^4$Genetics, $^5$Mathematics and Statistics\}
        } 
        \IEEEauthorblockN{ 
            \small $^6$Mila - Quebec AI Institute, $^7$Université de Montréal $^8$Wu Tsai Institute, $^9$Yale University 
             \ \ \ \ \ \  $^*$Jointly Supervised
        } 
    }
    \vspace{-.8cm}
}

\maketitle

\begin{abstract} 
Neuroscience employs diverse neuroimaging techniques, each offering distinct insights into brain activity, from electrophysiological recordings such as EEG, which have high temporal resolution, to hemodynamic modalities such as fMRI, which have increased spatial precision. However, integrating these heterogeneous data sources remains a challenge, which limits a comprehensive understanding of brain function. We present the Spatiotemporal Alignment of Multimodal Brain Activity (SAMBA) framework, which bridges the spatial and temporal resolution gaps across modalities by learning a unified latent space free of modality-specific biases. SAMBA introduces a novel attention-based wavelet decomposition for spectral filtering of electrophysiological recordings, graph attention networks to model functional connectivity between functional brain units, and recurrent layers to capture temporal autocorrelations in brain signal. We show that the training of SAMBA, aside from achieving translation, also learns a rich representation of brain information processing. We showcase this classify external stimuli driving brain activity from the representation learned in hidden layers of SAMBA, paving the way for broad downstream applications in neuroscience research and clinical contexts.
 
\end{abstract}

\section{Introduction}\label{sec:intro}

Non-invasive techniques such as electroencephalography (EEG) and magnetoencephalography (MEG) provide high temporal resolution, capturing the rapid dynamics of neural activity. In contrast, hemodynamic methods, such as functional magnetic resonance imaging (fMRI), offer rich spatial resolution~\cite{zhu2023statistical}. As neuroscience advances towards more sophisticated models of cognition, integrating these diverse data types becomes increasingly critical~\cite{boly2016functional}. Successfully combining the complementary strengths of these modalities could offer a more comprehensive understanding of brain function, but this remains a challenging task.

While substantial progress has been made in utilizing multimodal data consisting of image stimuli and brain activity pairs -- particularly with Generative Adversarial Networks (GANs), transformers, and diffusion models to reconstruct images from brain activity~\cite{haxby2001distributed, dado2022hyperrealistic, shen2019deep, seeliger2018generative, ozcelik2022reconstruction, lin2022mind, ho2020denoising, rombach2022high} -- the same is not true for the integration of multiple brain imaging modalities. Most of the work in this area has focused on leveraging information from EEG to enhance the fidelity of fMRI signals~\cite{calhas2020eeg, calhas2023eeg, abreu2018eeg, liu2019convolutional}. These efforts, while valuable in improving fMRI’s localization and signal-to-noise ratio with temporally rich EEG signals, often fall short of addressing the more complex task of multimodal fusion and do not address the complexities of spatiotemporal upsampling and downsampling between modalities.

To bridge this gap, we propose a novel multi-modal neural network framework, Spatiotemporal Alignment of Multimodal Brain Activity (SAMBA), designed to generalize the translation between electrophysiological and hemodynamic signals. SAMBA addresses both spatial and temporal disparities through graph attention and wavelet-based modules. Our objectives are threefold: (1) to create a unified latent space that captures spatiotemporal dynamics without modality-specific biases, enabling its application across a broad set of downstream tasks, such as brain state classification, cognitive assessment, and diagnosis of neurological disorders; (2) to develop data-driven models of hemodynamic response and functional connectivity in the brain; and (3) to combine smaller unimodal datasets into larger multimodal cohorts, laying the groundwork for training foundational models. SAMBA incorporates (1) temporal upsampling and downsampling modules based on learnable hemodynamic response functions (HRFs) and attention-based wavelet decomposition for spectral filtering; (2) spatial upsampling and downsampling modules powered by graph attention networks (GATs) to model functional connectivity across brain regions; and (3) recurrent layers to capture autocorrelations in the temporal domain.

We demonstrate the efficacy of SAMBA in several key tasks. First, the framework enables precise translation between electrophysiological and hemodynamic modalities, allowing for accurate cross-modal mapping. We also perform ablation studies to confirm the essential roles of all SAMBA components in achieving these results. Next, we show that SAMBA’s unified latent representations can accurately classify scenes in a movie shown to the subjects during data acquisition, demonstrating that the translation task allows SAMBA to capture rich representations of cognitive activity. Finally, we also show that the wavelet decomposition module in SAMBA filters specific EEG/MEG frequencies during translation for denoising, while the learnable HRF module models heterogeneity in neurovascular coupling across brain regions.

\begin{figure*}[htp]
    \centering   
    \includegraphics[width=.99\textwidth]{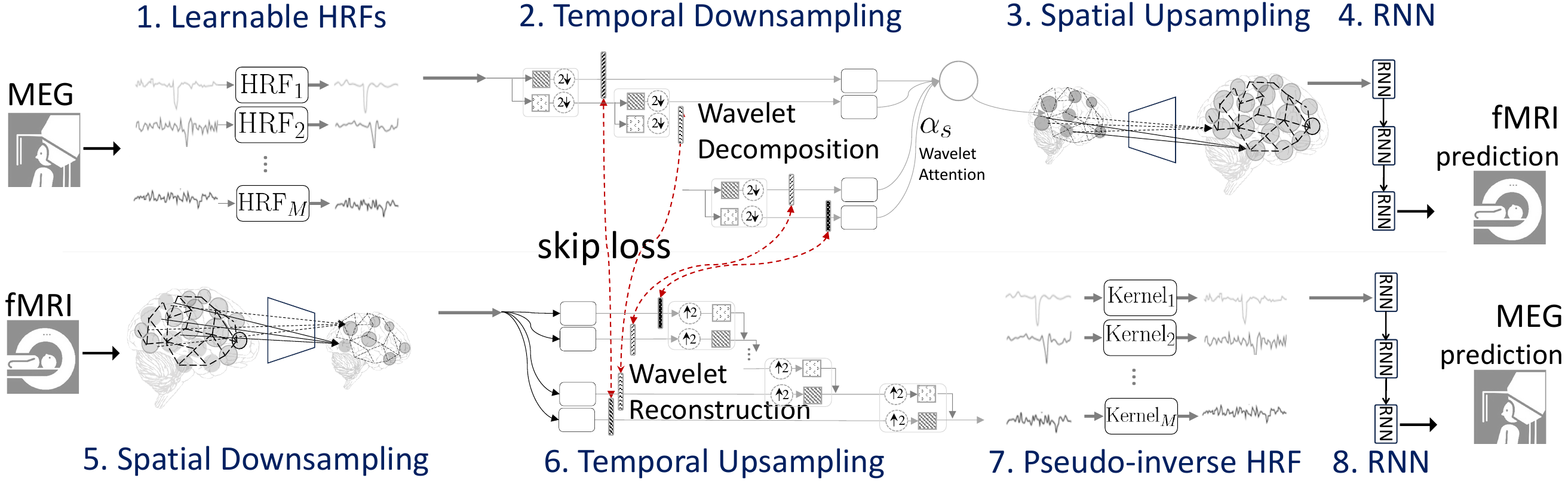}    
    \caption{\footnotesize SAMBA translates between MEG and fMRI modalities by upsampling and downsampling using wavelet decomposition and graph-attention modules in the temporal and spatial domains respectively. The upper and bottom parts show the fMRI-to-MEG and MEG-to-fMRI prediction modules respectively.
    }
    \label{fig: samba_overview}
    \vspace{-0.5cm}
\end{figure*}

\vspace{-.22cm}

\section{Methods} 
\label{sec:methods} 

Electrophysiological recordings, denoted as $X(t) = \{x_1(t), \ldots, x_N(t)\}$, represent the neural activity across $N$ parcels of the brain. Hemodynamic responses, represented as $Y(\tau) = \{y_1(\tau), \ldots, y_M(\tau)\}$, capture the blood oxygenation and flow changes across $M$ parcels, where $M \gg N$ due to the finer spatial resolution offered by fMRI. However, the temporal resolution of $X$ is higher than that of $Y$.

\subsection{Electrophysiological Activity to Hemodynamic Response} 
We elaborate on the translation from $X(t)$ to $Y(\tau)$.

\subsubsection{Temporal Smoothing with HRF learning} 

The HRF is designed to model the latency and variability of blood flow in response to neural activity. 
Due to significant variations in neuronal density and metabolic demand across regions of the brain, the HRF responses also vary across the brain \cite{herculano2009human,attwell2010glial}. To address this, we employ a parcel-specific HRF$_n (t)$, parameterized by six learnable parameters $\text{HRF}_n(t) =$
\begin{align}   
   \resizebox{0.44\textwidth}{!}{%
   $ \theta_{1} \left( \frac{t}{p_r} \right)^{\theta_{2}} \exp\left(-\frac{t - p_r}{\theta_{3}} \right)  
     - \theta_{4} \left( \frac{t}{p_u} \right)^{\theta_{5}} \exp\left(-\frac{t - p_u}{\theta_{6}} \right)$  
   }
\end{align}
where $\theta_{1}$ and $\theta_{4}$ are the amplitude of the response and undershoot components, respectively, modulating the increase and decrease in blood flow and oxygenation to the brain area activated following neural activity. $\theta_{2}$ and $\theta_{5}$ represent the time-to-peak of the response and undershoot components, respectively. $\theta_{3}$ and $\theta_{6}$ are the dispersion factors, influencing the width of the response and undershoot curves. $p_r = (\theta_{2} \cdot \theta_{3})$, and $p_u = (\theta_{5} \cdot \theta_{6})$ denote the peak times of the respective components.
The learnable parameters of the HRF model are inferred via a three-layer MLP for each brain parcel. For each parcel $n$, the HRF is convolved with the electrophysiological signal $x_n(t)$ to produce: $\tilde{x}_n(t) = \text{HRF}_n(t) \ast x_n(t)$,
where $\ast$ denotes the convolution operation. This convolution process smooths the electrophysiological signal into a representation of the blood flow dynamics resulting from neural activity.

\subsubsection{Temporal Downsampling}
\label{sec: temporal_downsample}

To perform temporal downsampling, we propose a unique architecture that compresses temporal signals via a rich wavelet transform and then uses attention to select the appropriate signal bands for the translation tasks. 
The process involves constructing daughter wavelets by scaling and translating the mother wavelet, $\psi$, by $s$ and $u$ respectively: $\psi_{s,u}(t) = \psi\left((t - u)/{s}\right)$.
Wavelet coefficients are computed by convolving $\tilde{x}_n(t)$ with daughter wavelets $c_n(s, u) = \tilde{{x}}_n(t) \ast \psi_{s, u}(t)$.
At smaller scales, where higher frequencies are analyzed, more translations $u$ are required to perform the convolution, resulting in a larger number of coefficients. Conversely, fewer translations are necessary at larger scales, yielding fewer coefficients.  
Next, we concatenate the scale-specific embeddings to form a multiscale representation, expressed as $z_n  = \parallel_s \alpha_s c_n(s)$. Here, $\quad z_n \in \mathbb{R}^d$ and $\alpha_s$ represents the learnable attention weight allocated to the embedding at scale $s$ normalized by the Softmax function, indicating the significance of features captured at that scale relative to others in the final multiscale representation. The attention weights are normalized using the Softmax function, transforming them into a probabilistic distribution that identifies the most salient frequency bands in the electrophysiological data.  
\begin{table*}[t]
    \centering
      \caption{\small Evaluation of translation using Spearman correlation for minute and second predictions between fMRI, MEG, and EEG modalities.
      }
      \label{tab: translation}
    \begin{subtable}{0.50\linewidth}
        \centering
        \label{tab:table1}
        \resizebox{\columnwidth}{!}{
        \begin{tabular}{rl!{\vrule}cccc }  
        \cmidrule(lr){2-6} 
        &  
        & \multicolumn{2}{c}{\textbf{MEG} $\rightarrow$ \textbf{fMRI}}  
        & \multicolumn{2}{c}{\textbf{EEG}$\rightarrow$\textbf{fMRI}}  
        \\  
        &                
        & {\footnotesize Minute} & {\footnotesize Second} 
        & {\footnotesize Minute} & {\footnotesize Second}   
        \\ 
        \cmidrule(lr){3-6}    
         \multirow{-2}{*}{\rotatebox{90}{a) \colorbox{lightorange}{Electrophysiological to Hemodynamic}}}  
                     
                     & MLP    
                     & 0.05   & 0.12  
                     & 0.04   & 0.10 
                     \\   
                     
                     & 1D-CNN  
                     & 0.07   & 0.14
                     & 0.09   & 0.14    
                     \\   
                     
                     & LSTM   
                     & 0.26   &0.39   
                     & 0.18   &0.31   
                     \\

                     & Transformer 
                     & 0.34    & 0.60   
                     & 0.19    & 0.28 
                     \\*[.5em]

                     & No Wavelet 
                     & 0.14   & 0.27      
                     & 0.12   & 0.24   
                     \\

                    & No LSTM 
                     & 0.18   & 0.30     
                     & 0.11   & 0.23  
                     \\   
                    
                     & No LSTM: Avg. 2 samples
                     & 0.36   & 0.65  
                     & 0.26   & 0.36  
                     \\  
        
                     & HRF-Wavelet-MLP-LSTM
                     & 0.37   & 0.66   
                     & 0.28   & 0.39 
                     \\  
                     
                     & Transformer instead of LSTM 
                     & 0.33   & 0.63      
                     & 0.23   & 0.37  
                     \\  
                     
                     & Fixed HRF    
                     & 0.36   & 0.60    
                     & 0.28   & 0.41     
                     \\ 
        
                     & MSE-Loss instead of Cosine 
                     & 0.36   & 0.58   
                     & 0.25   & 0.41 
                     \\
 
                     & SAMBA
                     & \textbf{0.38}   & \textbf{0.63}   
                     & \textbf{0.29}   & \textbf{0.43}   
                    \\ 
                    \cmidrule(lr){2-6} 
                     & {\color{blue}Transformer}
                     & {\color{blue}0.33}    & {\color{blue}0.62}    
                     & {\color{blue}0.14}    & {\color{blue}0.30}  
                     \\ 
                    
                    & {\color{blue}SAMBA}
                    & {\color{blue}\textbf{0.39}}    & {\color{blue}\textbf{0.67}}   
                    & {\color{blue}\textbf{0.28}}    & {\color{blue}\textbf{0.44}}   
                    \\
        \cmidrule(lr){2-6} 
        \end{tabular}
        } 
    \end{subtable}%
    \begin{subtable}{0.47\textwidth}
        \centering 
        
        \label{tab:table2}
        \resizebox{\columnwidth}{!}{
        \begin{tabular}{rl!{\vrule}cccc}
            \cmidrule(lr){2-6}  
            &   
            & \multicolumn{2}{c}{\textbf{fMRI}$\rightarrow$\textbf{MEG}}   
            & \multicolumn{2}{c}{\textbf{fMRI}$\rightarrow$\textbf{EEG}}   
            \\  
            &                
            & {\footnotesize Minute} & {\footnotesize Second} 
            & {\footnotesize Minute} & {\footnotesize Second}  
            \\ 
            \cmidrule(lr){3-6}  

        \multirow{-3}{*}{\rotatebox{90}{b) \colorbox{lightyellow}{ Hemodynamic to Electro.}}} 
                     
                     & MLP     
                     & 0.05   & 0.11  
                     & 0.05   & 0.10
                     \\
                     
                     & 1D-CNN   
                     & 0.06   & 0.16     
                     & 0.07   & 0.15 
                     \\
                     
                     & LSTM    
                     & 0.13   &0.25   
                     & 0.11   &0.22 
                     \\

                     & Transformer  
                     & 0.15    & 0.30   
                     & 0.11    & 0.28  
                     \\
  
                     & No Pseud HRF     
                     & 0.21   & 0.34  
                     & 0.09   & 0.20 
                     \\
                     
                     & No Skip Loss    
                     & 0.15   & 0.24   
                     & 0.10   & 0.19  
                     \\ 

                     & SAMBA 
                     & \textbf{0.21}   & \textbf{0.35} 
                     & \textbf{0.15}   & \textbf{0.33}   
                    \\ 
                     \cmidrule(lr){2-6} 
                     & {\color{blue}Transformer} 
                     & {\color{blue}0.11}    & {\color{blue}0.26}   
                     & {\color{blue}0.10}    & {\color{blue}0.26}  
                     \\   
                    
                    & {\color{blue}SAMBA} 
                    & {\color{blue}\textbf{0.19}}    & {\color{blue}\textbf{0.31}}   
                    & {\color{blue}\textbf{0.13}}    & {\color{blue}\textbf{0.27}}  
                    \\
        \cmidrule(lr){2-6} 
        \end{tabular}
        } 
        
        {\ \\  \color{blue} We built a baseline using five methods between withheld time intervals for all subjects and subject pairs (blue-coded).}
    \end{subtable}
    \vspace{-0.5cm}
\end{table*}

\subsubsection{Spatial Upsampling Module}

In this module, we outline our approach for translating data from a coarse-grained graph of brain regions, denoted as $G_X = (V^X, E^X, W^X)$, derived from electrophysiological measurements in the source modality, to a fine-grained graph, $G_Y = (V^Y, E^Y, W^Y)$, which features a higher spatial resolution using hemodynamic data from the target modality (Fig.~\ref{fig: samba_overview}c). Recall that our task is to translate $N$ time-lapse electrophysiological signals represented as $X(t) = \{ x_1(t), \ldots, x_N(t) \}$, to $M$ time-lapse hemodynamic signals $Y(\tau) = \{ y_1(\tau), \ldots, y_M(\tau) \}$. To achieve this, our source graph contains $N$ nodes ($|V_X| = N$) and our target graph contains $M$ nodes ($|V_Y| = M$), where $M \gg N$.
Here, the edge weights, $W^X$, in the source graph, are assigned based on the cosine similarity between timelapse electrophysiological signals: $W^X_{pq} = (x_p(t) - \bar{x}_p) \cdot (x_q(t) - \bar{x}_q)/\|x_p(t) - \bar{x}_p\| \|x_q(t) - \bar{x}_q\|,$
where $\bar{x}_p$ is the mean of the signal $x_p(t)$.  
We input the latent representations $\{z_j\}_{j=1}^N$ as node features into a GAT layer, which computes hidden features of nodes  
\begin{equation}  
    \label{eq:src_gat_update}
    \resizebox{0.42\textwidth}{!}{%
    $h_n^{X}(\tau) =  \sigma\left(\frac{1}{K} \sum_{k=1}^K \sum_{j \in \mathcal{N}(n)} \beta_{nj}^{(k)} {W}^{(k)} {z}_j(\tau) \right),$
    }
\end{equation}
where $K$ is the number of attention heads, $\beta^{(k)}$ are the attention coefficients, and $W^{(k)}$ are the head-specific weight matrices. We then follow the standard GAT implementation~\cite{velivckovic2017graph, brody2021attentive}. Edge weights in the target graph $G_Y$ are based on the cosine similarity of hemodynamic signals: $W^Y_{pq} = (y_p(\tau) - \bar{y}_p) \cdot (y_q(\tau) - \bar{y}_q)/\|y_p(\tau) - \bar{y}_p\| \|y_q(\tau) - \bar{y}_q\|$
where, $\bar{y}_p$ denotes the mean of the hemodynamic signal in parcel $p$.  
The node features in $G_Y$ are defined using single-layer feed-forward networks, $\{\phi_m\}_{m=1}^M$, which map the hidden representations $\{h^X_n\}_{n=1}^N$ in $G_X$ to the nodes in $G_Y$. Each network $\phi_m$ takes the aggregated representations $\{ h_i^X \}_{i \in \chi_m}$ as input, where $\chi_m$ is the subset of nodes from the same neuroanatomical region in the source graph. For example, to obtain the node features of a visual cortex parcel in the target graph, $G_Y$, we use hidden representations of all available visual cortex parcels in the source graph, $G_X$. 
We then used a GAT layer to aggregate the features in the target graph: 
\begin{equation}  
    \label{eq:tgt_gat_update}
    \resizebox{0.44\textwidth}{!}{%
    $h_m^{Y} (\tau) =  \sigma\left(\frac{1}{K} \sum_{k=1}^K \sum_{j \in \mathcal{N}(m)} \gamma_{mj}^{(k)} {W}^{(k)} \phi_m(\{ h_i^X \}_{i \in \chi_m}) \right),$ 
    }
\end{equation}
where, $\gamma^{(k)}$ are normalized attention coefficients, $\mathcal{N}(m)$ is neighboring nodes of $m$, and $W^{(k)}$ are unique weight matrices for each attention head.
Ultimately, this module generates a series of high-resolution node representations, $\{h^Y_m\}_{m=1}^M$, which produce the desired output, $Y(\tau)$.  

\subsubsection{Hemodynamic Sequence Generation via RNNs}

Upon spatially upscaling, the refined high-resolution node representations, denoted as $h_m^{Y}$, are fed into a recurrent model in the final stage.  
To this end, we employ a LSTM network, since it is well-suited for modeling the autoregressive characteristics inherent in these temporal sequences. The LSTM processes the sequence of node representations, $h_m^{Y}(\tau)$, to predict hemodynamic activity, $\hat{Y}(\tau) = \{ \hat{y_1}(\tau), \cdots, \hat{y}_M(\tau) \}$, as follows:
\begin{align}\label{eq: lstm} 
\hat{y}_m(\tau_o + 1) = \text{LSTM}(\hat{y}_m(\tau_o), h_m^{Y}( \tau_o + 1)),  
\end{align}
where $ m = 1, \cdots M$ and $\hat{y}_m(\tau_o+1)$ is the estimated hemodynamic activity in the $m$-th parcel at time $\tau = \tau_o + 1$.  This estimation relies on the previously predicted $\tau_o$, denoted as $\hat{y}_m(\tau_o)$, and the current node representation, $h_m^{Y}( \tau_o + 1)$.

\subsection{Hemodynamic Response to Electrophysiological Activity}
Here, we describe our methodology for converting hemodynamic activity, $Y(\tau)$, to electrophysiological activity, $X(t)$.  
 
\subsubsection{Spatial Downsampling Module}   

To perform spatial downsampling, we invert and adapt the methodology detailed in the graph upsampling section, converting a fine-grained hemodynamic graph, $G_Y$, containing $M$ nodes, into a coarse-grained electrophysiological graph, $G_X$, containing $N$ nodes, where $M \gg N$. Here, a GAT layer aggregates node features from the brain activity graph $G_Y$, which are then mapped to a coarser target graph $G_X$ using linear layers. 
 
\begin{figure*}[h]
    \centering
    \begin{minipage}[t]{0.64\textwidth}
        \centering
        \vspace{-10pt}
         \includegraphics[width=0.99\linewidth]{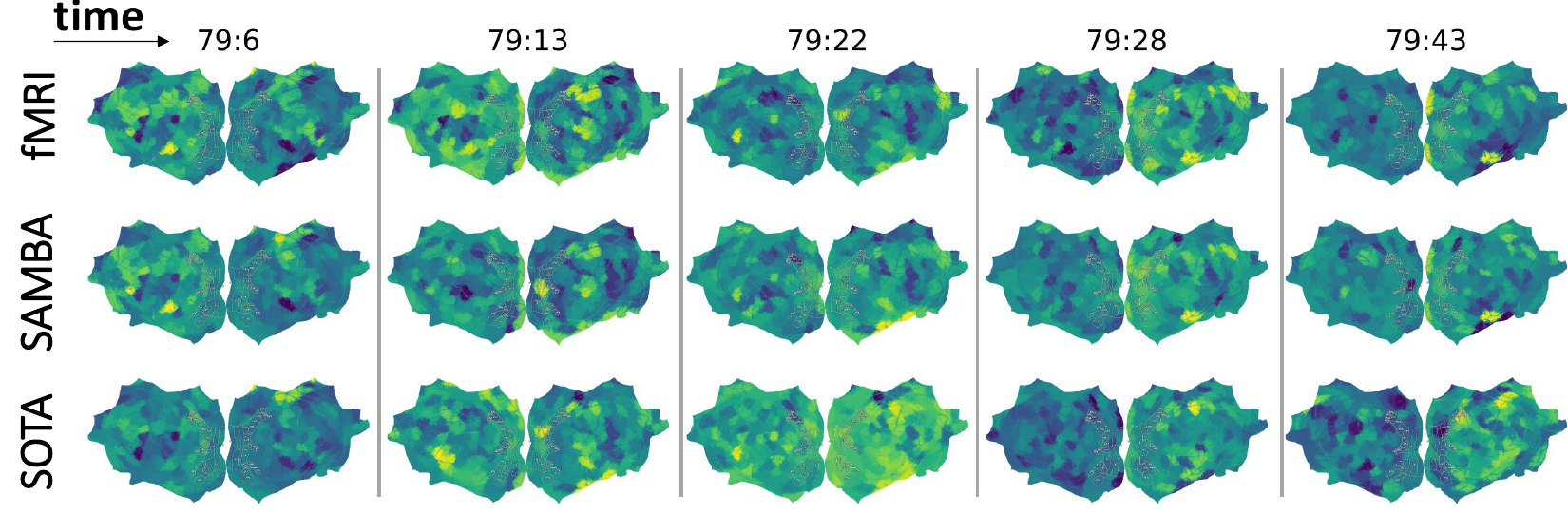}
         \vspace{-.2cm}
       \caption{\footnotesize PyCortex~\cite{gao2015pycortex} visualizations of fMRI activity on the unfolded brain surface, comparing ground truth (first row) with translations obtained via SAMBA (middle row) and the SOTA transformer model (third row). Timestamps (mm:ss) in columns correspond to the Forrest Gump movie.}
        \vspace{-1cm}
        \label{fig: reconstruction}
    \end{minipage}\hfill
    \begin{minipage}[t]{0.34\textwidth}
        \centering
        \vspace{-10pt} 
        \includegraphics[width=\textwidth]{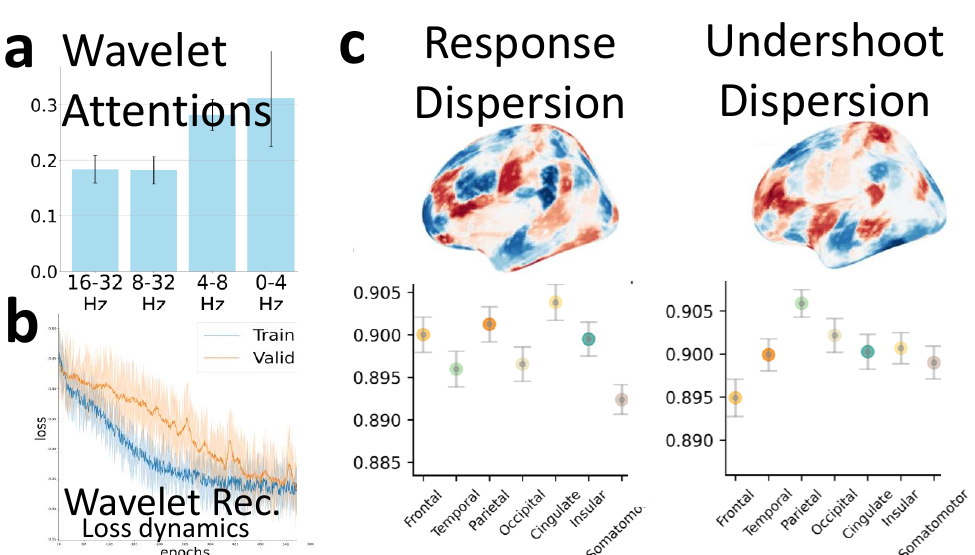}
        \caption{\footnotesize Wavelet attention in a, and reconstruction loss dynamics in b. c) Inferred HRF undershoot and response dispersion parameters.
        } 
        \label{fig: wavelet_and_hrf_params}
    \end{minipage}\hfill 
     \vspace{-14pt} 
\end{figure*}

\subsubsection{Temporal Upsampling Module} 
Given $h_n^{X} (\tau)$, as the spatially downsampled hemodynamic data, we now aim to perform temporal upsampling.  We first model the reverse process of wavelet decomposition by estimating the wavelet coefficients at various wavelet coefficient scales and performing the inverse wavelet decomposition. We achieve this in two steps. First, we estimate the wavelet coefficients using a set of linear layers $\{f_s\}_{s=1}^{\mathcal{S}}.$ Each layer $f_s$ maps the input signal to the wavelet coefficient space at a specific scale: $\hat{c}(s, u) = f_s(h_n^{X} (\tau)),$
where $\hat{c}(s, u)$ represents the estimated wavelet coefficient at scale $s$ and position $u$. To reconstruct $n$-th HRF smoothed signal, we then perform wavelet reconstruction using the estimated coefficients:
\begin{equation}
    \resizebox{0.32\textwidth}{!}{%
  $\tilde{{x}}_n(t) = \sum_{s \in \mathcal{S}} \sum_{u \in \mathcal{U}} \hat{c}(s, u) \psi_{s,u}(t),$
  }
\end{equation}

where $\psi_{s,u}(t)$ denotes the daughter wavelets obtained by scaling and translating the mother wavelet $\psi$ by factors of $s$ and $u$, respectively. However, to ensure accurate wavelet coefficient estimation, we employ a regularization strategy using wavelet coefficient skip losses (between blocks 1 and 6 in Fig.~\ref{fig: samba_overview}). This function penalizes the network for discrepancies between the true wavelet coefficients $c(s, u)$ from the electrophysiological, and the estimated coefficients $\hat{c}(s, u)$:
\begin{equation}
\label{loss: reg_loss}
\resizebox{0.4\textwidth}{!}{%
  ${L}_{\text{reg}} = \frac{1}{|\mathcal{S}|} \frac{1}{|\mathcal{U}|} \sum_{s \in \mathcal{S}}  \sum_{u \in \mathcal{U}}  (c(s, u) - \hat{c}(s, u))^2.$
  }
\end{equation}

\subsubsection{Deconvolution using Pseudo-inverse HRF}
 
We now aim to build a pseudo-inverse HRF function to estimate the original neural signals from smoothed HRF. 
Since the double gamma form of the HRF function is not invertible, we estimate the original temporal dimension of MEG or EEG (at 200 Hz) using per-parcel single kernel learning via 1D transpose convolution. The reconstruction is mathematically represented as: $\hat{{x}}_n(t) = \text{DeConv1D}_n (\tilde{{x}}_n(t)), $
where $\text{DeConv1D}_n$ is the parcel-specific transpose convolution with the single learnable kernel.

\subsubsection{Electrophysiological Sequence Generation with RNNs} 
Upon temporal reconstruction, the refined low-resolution node representations, denoted as \(h_n^{X}\), are fed into a recurrent model in the final stage of translation from hemodynamic activity to electrophysiological signals in the brain.  
To this end, we employ an LSTM to process the sequence of node representations, \(h_n^{X}(t)\), to predict electrophysiological activity, $\hat{X}(t) = \{ \hat{x_1}(t), \cdots, \hat{x}_N(t) \}$, akin to Eq.~\ref{eq: lstm}.

\subsection{Loss Formulation}
We employed the cosine similarity loss function to train the model to align the predicted signal with the target signal. 
In hemodynamic mapping to electrophysiological, for example, given the predicted $m$-th parcel $\hat{{y}}_m$, the loss is defined as:
\begin{equation}\label{loss: cos}
    \resizebox{0.35\textwidth}{!}{%
    ${L}_{\text{match}}  = \sum_{m=1}^M \big( 1- \frac{\hat{{y}}_m  \cdot {y}_m}{\|\hat{{y}}_m \|_2 \, \|{y}_m\|_2} \big),$
    }
\end{equation} 
where, $M$ is the number of parcels, \( \|\hat{{y}}_m \|_2 \), and \( \|{y}_m\|_2 \) are the L2 norms of \(\hat{{y}}_m\) and \( {y}_m \), respectively. Here, in addition to the cosine loss we also regularized the network using skip loss, as in Eq.~\ref{loss: reg_loss}: $\lambda {L}_{\text{match}} + (1-\lambda){L}_{\text{reg}}$. However, to map electrophysiological to hemodynamics we only train the model with the cosine loss Eq.~\ref{loss: cos}, given $\hat{{y}}_n$ and $y_n$.

\section{Results}
\label{sec: evaluation}

We conduct experiments using two datasets: (1) StudyForrest~\cite{hanke2016studyforrest, liu2022studyforrest}, which comprises MEG and fMRI data, and (2) Naturalistic Viewing~\cite{telesford2023open}, which includes EEG and fMRI recordings. To this end, we evaluate SAMBA
on four translation tasks: (1) fMRI-to-MEG, (2) fMRI-to-EEG, (3) MEG-to-fMRI, and (4) EEG-to-fMRI. We then explore our SAMBA model's evaluation of the classification task to detect eight distinct movies in the Naturalistic Viewing dataset.

In Table~\ref{tab: translation}, we compare SAMBA's performance against several baseline architectures, including convolutional, transformer, recurrent, and feed-forward networks. We also include ablation studies of the SAMBA architecture, where key components such as wavelet decomposition, the learnable HRF, and the recurrent layer are systematically removed or replaced. Specifically, in Table~\ref{tab: translation}a we assess performance in translating electrophysiological data to hemodynamic data, and in Table ~\ref{tab: translation}b, we report results for the reverse task. The primary evaluation metric is Spearman correlation, averaged across all Schaefer parcels, between the predicted and ground truth time-lapse signals in both long (1 min) and short (15 sec) intervals of withheld timepoints. The evaluate SAMBA when trained across all fMRI-EEG/MEG subject pairs (black text), as well as a subject-specific SAMBA model, where a separate model is trained for each subject pair (blue text), and the reported Spearman correlations are averaged across all withheld timepoints for each subject. SAMBA outperforms all baseline models across all tasks, with the transformer model by Vaswani et al.~\cite{vaswani2017attention} achieving the second-best performance.

Fig.~\ref{fig: reconstruction} illustrates SAMBA's performance in translating MEG to fMRI data, using pycortex\cite{gao2015pycortex} from the StudyForest dataset. While the first row presents ground-truth fMRI recordings, the second and third rows show SAMBA and SOTA (transformer) reconstructions over the brain surface. The results indicate that SAMBA effectively recovers fMRI signals from MEG measurements, even in the test set.

Fig.~\ref{fig: wavelet_and_hrf_params}a illustrates the dynamics of wavelet decomposition attention and wavelet reconstruction skip loss in our model. Based on the attention intensity values, our models primarily focus on lower frequencies (4-8 Hz and 0-4 Hz), likely due to the higher signal-to-noise ratio at these frequencies compared to higher frequencies. Fig.~\ref{fig: wavelet_and_hrf_params}b presents variations in the details of the skip-loss dynamics during wavelet reconstruction.

To showcase the richness of the representation learned by SAMBA, we added a classification head to identify eight distinct movies from the Naturalistic Viewing dataset~\cite{telesford2023open}. Table~\ref{tab: classification} compares our model's performance against baseline methods. Notably, our model achieves a 10.54\% improvement in the EEG to fMRI classification tasks over the baseline.
\begin{table}[ht]
    \centering
    \captionof{table}{  Movie classification accuracy results.}
    \vspace{-0.1cm}
    \begin{tabular}{rl!{\vrule}cc}
                \toprule       
                &              
                & EEG-to-fMRI & fMRI-to-EEG   
                 \\   
                \cmidrule(lr){1-4}    
                 & 1D-CNN 
                 & 48.83\tperc   & 30.69\tperc      
                 \\   
                 & LSTM 
                 & 53.71\tperc   & 37.09\tperc      
                 \\   
                 & Transformer
                 & 51.04\tperc   & 38.24\tperc        
                 \\
                 & SAMBA
                 & \textbf{61.58}\tperc  & \textbf{46.50}\tperc              
                  \\     
                \bottomrule   
                
            \end{tabular}
    \label{tab: classification}
    \vspace{-0.4cm}
\end{table}

Our model also offers neuroscientific interpretations. Here, we outline key findings from the best-performing MEG-to-fMRI model. Fig.~\ref{fig: wavelet_and_hrf_params}c displays the inferred HRF parameters for each brain parcel. This figure shows the variation in HRF response and undershoot dispersion across different brain regions, highlighting the diversity in oxygenation and deoxygenation levels~\cite{herculano2009human}. Notably, the left somatomotor network exhibits minimal response dispersion compared to the cingulate, whereas the parietal lobe regions show greater undershoot dispersion than those in the right somatomotor network. 
   
\section{Conclusions}
\label{sec: conc_limits}

This paper introduces SAMBA, a framework designed to address spatiotemporal trade-offs in multimodal brain activity translation. Using wavelet-attention-based temporal encoding and decoding with context-aware graph upsampling and downsampling, SAMBA outperforms baseline methods like transformers. The framework's translation task yields rich representations useful for downstream tasks like classification.

{\small
    \bibliographystyle{ieee_fullname}
    \bibliography{egbib.bib}
}

\end{document}